\documentclass[11pt]{article}
\pdfoutput=1
\usepackage[utf8]{inputenc}
\usepackage[OT1]{fontenc}
\usepackage[usenames]{color}
\usepackage{url}            
\usepackage{booktabs} 
\usepackage{amsmath}      
\usepackage{amsthm}
\usepackage{amsfonts}       
\usepackage{nicefrac}       
\usepackage{microtype}      
\usepackage{xcolor}         
\usepackage{tablefootnote}
\newtheorem{definition}{Definition}

\usepackage{fullpage}

\usepackage[colorlinks,linkcolor=red,anchorcolor=blue,citecolor=blue]{hyperref}

\usepackage{smile}
\usepackage{cme-math}

\def\score{\mathop{\text{score}}}
\def\enc{\mathop{\text{encoder}}}
\def\dec{\mathop{\text{decoder}}}

\begin{document}

\title{MQRetNN: Multi-Horizon Time Series Forecasting with Retrieval Augmentation}
\author{Sitan Yang\thanks{Amazon Forecasting Science, Correspondence to: sitanyan@amazon.com}
	\and Carson Eisenach\footnotemark[1]
	\and Dhruv Madeka\footnotemark[1]}

\maketitle

\begin{abstract}
	Multi-horizon probabilistic time series forecasting has wide applicability to real-world tasks such as demand forecasting. Recent work in neural time-series forecasting mainly focus on the use of Seq2Seq architectures \cite{Sutskever2014}. For example, MQTransformer \citep{Eisenach2020} -- an improvement of MQCNN \citep{Wen2017} -- has shown the state-of-the-art performance in probabilistic demand forecasting. In this paper, we consider incorporating cross-entity information to enhance model performance by adding a cross-entity attention mechanism along with a retrieval mechanism to select which entities to attend over. We demonstrate how our new neural architecture, MQRetNN, leverages the encoded contexts from a pretrained baseline model on the entire population to improve forecasting accuracy. Using MQCNN as the baseline model (due to computational constraints, we do not use MQTransformer), we first show on a small demand forecasting dataset that it is possible to achieve $\sim$3\% improvement in test loss by adding a cross-entity attention mechanism where each entity attends to all others in the population. We then evaluate the model with our proposed retrieval methods -- as a means of approximating an attention over a large population -- on a large-scale demand forecasting application with over 2 million products and observe $\sim$1\% performance gain over the MQCNN baseline.
\end{abstract}

\section{Introduction}
Multi-horizon probabilistic time series forecasting has many important applications in real-world tasks \cite{Carlos2016,Wen2017, Bose2017, Lim2018, madeka2018sample, Eisenach2020}.  For example,
consider a retailer who wishes to optimize their purchasing decisions. In order to make optimal decisions, they require forecasts of consumer demand at multiple time steps in the future. In the domain of multi-horizon, probabilistic time-series forecasting, deep neural networks (DNNs), especially those of the Seq2Seq variety \cite{Sutskever2014}, have increasingly been studied recently \cite{Wen2017, Flunkert2017, Lim2019, Eisenach2020, olivares2021probabilistic}. They have various advantages over traditional time series models including the ability to easily handle a complex mix of historic covariates and the potential to incorporate recent advances in Seq2Seq learning.

Like many other machine learning tasks, the canonical formulation of a forecasting model in this case considers time series for $N$ entities -- e.g. $N$ products in the case of demand forecasting -- and forecasts are produced using features specific to that entity. Models are trained using shared weights for each entity. Seq2Seq architectures consist of an \emph{encoder}, which typically summarizes time-series covariates for an entity $i$ into time specific representations, whch we denote as $h_{i,t}$ and a {\it decoder} which takes the encoded context and produces the output sequence (in this case, probabilistic forecasts). For an entity $i$ and time $t$, we expect $h_{i,t}$ to be more relevant to the forecast target in inference than $h_{j,t}$ for any other $j \neq i$. That does not mean, however, that the encoded contexts of other entities contain no relevant information. As an example, consider forecasting demand for soda from two competing brands (brand A and brand B) -- these products may be substitutes, and when the demand goes down for one, the other increases. In this way, information from brand A may be useful for forecasting for brand B, and vice-versa. Traditional Seq2Seq neural architectures such as RNN (Recurrent Neural Network), LSTM (Long Short-Term Memory network) and CNN (Convolution Neural Network) fail to capture such {\it cross-entity information}.

In natural language processing (NLP), a recent advance is retrieval-based language models \cite{kelvin2020, urvashi2020, Borgeaud2021} that directly search and utilize information from a large corpus, such as Wikipedia, to help inform predictions. For example, Retrieval-Enhanced Transformer (RETRO) introduced in \cite{Borgeaud2021} imprtoves language model performance not by scaling up model parameters or training data size, but via learning on information retrieved from a task-related database. The key idea is to apply attention \cite{Bahdanau2014} over representations of other entities in the population. Because the size of the population can be quite large, the authors propose using a $k$-nearest neighbors ($k$-NN) search similar to \cite{urvashi2020} to lookup up the most relevant entities and attend only over those.

Inspired by these studies, we introduce a cross-entity attention mechanism along with a retrieval mechanism to the state-of-the art MQ-Forecaster framework \cite{Wen2017, Wen2019, Eisenach2020} for probabilistic time series forecasting. In particular, we build this work on the MQCNN model \cite{Wen2017} but our methods can be naturally extended to any generic Seq2Seq time series forecaster. For retrieval methods, in addition to the commonly used $k$-NN search, we also propose using an arbitrary submodular function to select a relevant set of entities to attend over and motivate the use of a submodular function to approximate an attention mechanism.

Our work is one of the first architectures to leverage cross-entity information with retrieval-augmentation, and to the best of our knowledge, is the first to do so in the domain of time-series forecasting. Our main contributions are the following:
\begin{enumerate}
	\item MQRetNN -- a retrieval-based Seq2Seq architecture for multi-horizon time series forecasting. The model builds on the encoder-decoder architecture of MQCNN, and utilizes an offline database of entity representations which the model attends over during training and inference. We also incorporate a retrieval mechanism to efficiently select which entities to attend over so that the methodology can scale to large datasets. We show that our model brings noticeable accuracy gains over the MQCNN baseline on both a small-scale and large-scale demand forecasting problem.
	
	\item A new retrieval method that uses a submodular scoring function to efficiently summarize all contexts from the offline database rather than searching for nearest neighbors of each example as commonly used in the literature. As we show in the results section, this method achieves comparable performance to the $k$-NN search in our applications.
\end{enumerate}

The rest of the paper is organized as follows: in Section \ref{sec:background}, we provide an overview of the multi-horizon time series forecasting problem and related work. In Section \ref{sec:method} we describe our proposed methods in detail. In Section \ref{sec:results} we present the experimental results. We show that on our target application -- demand forecasting -- it is possible to achieve a 3\% improvement over the baseline when attending over all other entities in the population on a small dataset (approximately 10K products). We then evaluate several retrieval mechanisms that scale the model to a much larger population (around 2M products) and allow us to obtain an improvement of approximately 1\% over the baseline.

\section{Background and Related Work}
\label{sec:background}

\subsection{Time-Series Forecasting}
We consider the high-dimensional regression problem with a mix of inputs where at each time $t$ and for each entity $i$, we forecast the distribution of $y$ over the next $H$ periods:
\begin{align}
\label{eq:basic_problem1}
p\left(y_{i, t+1}, \ldots, y_{i, t + H} | y_{i, :t}, x_{i, :t}^{(h)}, x_{i, t:}^{(f)}, x_{i}^{(s)} \right),
\end{align}
where $y_{i, \cdot}$ denotes the target time series of entity $i$, $x_{i, :t}^{(h)}$ are historic covariates up through time $t$, $x_{i, t:}^{(f)}$ are covariates that are known apriori (such as calendar information), and $x_{i}^{(s)}$ are static covariates.

Many recent works \cite{Wen2017, Lim2019, Eisenach2020} have considered this forecasting problem. In this paper, our application of interest is demand forecasting for a large e-commerce retailer and downstream applications require only specific quantiles, not the full distribution. Accordingly, we focus on producing quantile forecasts similar to other recent works \cite{Wen2017,Lim2019,Eisenach2020}. Our model architecture builds off of the MQCNN architecture introduced in \cite{Wen2017}.

\subsection{Attention Mechanisms}
Attention mechanisms \cite{Bahdanau2014, Galassi2019} compute an alignment between a set of {\it queries} and {\it keys} to extract a {\it value}. Formally, let $\qb_1,\dots,\qb_t$, $\kb_1,\dots,\kb_t$ and $\vb_1,\dots,\vb_t$ be a series of queries, keys and values, respectively. The $s^{th}$ {\it attended value} is defined as
\[
\cbb_s = \sum_{i=1}^t \score(\qb_s,\kb_t)\vb_t,
\]
where $\score$ is a scoring function -- commonly $\score(\ub,\vb) := \ub^\top\vb$. Often, one takes  $\qb_s = \kb_s = \vb_s = \hb_s$, where $\hb_s$ is the hidden state at time $s$.

The transformer architecture was first proposed in \cite{Vaswani2017} and achieved state-of-the-art performance in language modeling. In the vanilla transformer, each encoder layer consists of a multi-headed attention block followed by a feed-forward sub-layer. For each head $i$, the attention score between query $\qb_s$ and key $\kb_t$ is defined as follows
\begin{equation}
\label{eqn:vanilla_attn}
A^h_{s,t} = \qb_s^\top \Wb_q^{h,\top} \Wb_k^{h}\kb_t.
\end{equation}
This architecture design has been successfully adopted in many subsequent studies with various extensions such as Transformer-XL \cite{Dai2019}, Reformer \cite{Nikita2020} and most recently Retrieval-Enhanced Transformer \cite{Borgeaud2021}.

\subsection{Retrieval Mechanisms}
Information retrieval is a classic topic for language modeling and a recent advance is the retrieval-based models. Several latest works have demonstrated the benefit of adding an explicit retrieval step to neural networks. In \cite{urvashi2020}, kNN-LM is proposed to enhance a language model through a nearest neighbor search in suitable text collections. \cite{kelvin2020} introduces REALM which augments language model pretraining with a latent knowledge retriever. More recently, RETRO \cite{Borgeaud2021} enhances the model architecture not by increasing the number of parameters or the size of training data, but rather through the retrieval of information relevant for each sample. Similarly, \cite{Bonetta2021} uses memorized similarity information from the training data for retrieval at at inference time.

\subsection{Data Summarization and Submodular Functions}
Data summarization has gained a lot of interest in recent years with the application of so called  \emph{Submodular Functions}. Applications range from exemplar-based clustering \cite{Dueck2007} to document summarization \cite{dasg2013, Lin2011}. The goal is to select representative subsets of elements from a large-scale dataset through a pre-defined optimization process. The key component of the optimization formulation is a submodular function which serves as a scoring function for any particular subset.

\begin{definition}[Submodular Function]
	\label{def:submodular}
	Let $\Omega$ be a finite set. A function $f: 2^\Omega \rightarrow \RR$ is said to be submodular if for any $S \subseteq T \subseteq \Omega$ and any $x \in \Omega \setminus S$
	\[
	f(S\cup\{x\}) - f(S) \geq f(T\cup\{x\}) - f(T).
	\]
\end{definition}

The essential property of submodular functions is known as submodularity, an intuitive diminishing returns condition that allows the search for nearly-optimal solutions in linear time and fits well into the purpose of subset selection. The formal definition is given as Definition \ref{def:submodular} and we direct the reader to \cite{Krause2014} for a thorough overview of submodular functions and their optimization. Many recent applications of submodular optimization focus on scaling up traditional algorithms to dealing with massive amounts of data or data streams. Proposed methods include distributed algorithms \cite{Mirz2013, Kumar15} and streaming algorithms \cite{Badan2014}.

\section{Methodology}
\label{sec:method}

\subsection{Problem Formulation}

As mentioned in Section \ref{sec:background}, we aim to estimate the distribution of the target variable $y_i$ as presented in Equation \eqref{eq:basic_problem1} over the next $H$ horizons at each time $t$. We train a quantile regression model to minimize the total \emph{quantile loss}, summed over all forecast creation times (FCTs) $T$ with $Q$ quantiles and $H$ horizons
\begin{equation}
\label{eqn:train_loss}
\sum_{t}\sum_{q}\sum_{h} L_q\left(y_{i, t+h}, \hat{y}_{i, t+h}^{(q)}\right),
\end{equation}
where $L_q(y,\hat{y}) = q(y-\hat{y})_{+} + (1-q)(\hat{y}-y)_{+}$, $(\cdot)_+$ is the positive part operator, $t$ denotes a FCT, $q$ denotes a quantile, and $h$ denotes the horizon. In this paper, we adopt the multi-horizon forecasting setting described in \cite{Wen2017, Lim2019, Eisenach2020} with the output of the 50th and 90th percentiles (P50 and P90) at each time step, and thus the model is trained to jointly minimize the P50 and P90 quantile loss.

\subsection{Model Architecture}
We design our model to be capable of leveraging an offline database constructed using the encoded representations from a frozen, pre-trained base model.
In this paper, we use MQCNN \cite{Wen2017} as the base architecture rather than the state-of-the-art MQTransformer \cite{Eisenach2020} as the latter one requires substantially more GPU memory and, as discussed below, we are already memory bound. Further, we expect that our retrieval mechanism offers an improvement that is orthogonal to those in MQTransformer, and the two sets of improvements could be combined in future work.

Generally our model adopts the Seq2Seq structure of MQCNN with an encoder that produces an encoded context at time $t$
\[
h_{i,t} := \enc(y_{i,:t},x_{i,:t}^{(h)},x^{(s)}_i),
\]
and a decoder that differs from MQCNN in that we include an additional ``cross-entity context'', which we denote as $\tilde{h}_{i,t}$. Formally, the decoder computes
\[
\hat{\Yb}_{i,t} := \dec(h_{i,t},\tilde{h}_{i,t},x_{i,t:}^{(f)})
\]
where $\hat{\Yb}_{i,t}$ is a matrix of shape $H\times Q$ for forecast quantiles of different horizons. We also denote $\Hb := \{ h_{i,t} | \forall i, t \}$ and $\tilde{\Hb} := \{ \tilde{h}_{i,t} | \forall i, t \}$. Ideally, $\tilde{\Hb}$ would be computed by attending \emph{all other} entities in the database, but for large datasets this may become infeasible. Thus we add a retrieval mechanism to select an informative subset of entities to attend across at each time step, which we provide more details in the next section.

To generate $\tilde{\Hb}$, we add a \emph{time series cross-attention} layer after the encoder to extract the cross-entity information through attention between the retrieved contexts and examples during training. The attention is computed only at each time step across different entities, and no cross time (temporal) attention is currently considered. Proper masking is used to make sure the attended and attending entities are aligned as shown in Figure \ref{fig:cross-att}.

We find in our experiments that this process increases the GPU memory consumption of the model because the retrieved contexts are loaded with each mini-batch during training. This in turn limits the total number of elements contained in these contexts, which makes the retrieval mechanism necessary on large datasets.

\begin{figure}[!h]
	\centering
	\includegraphics[scale=.6]{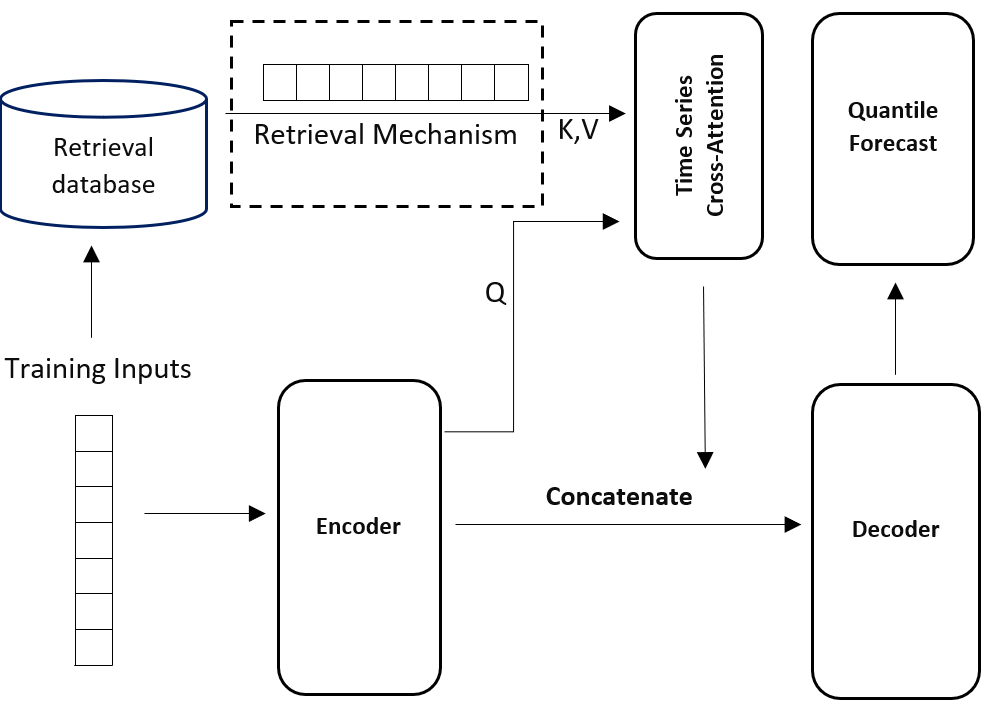}
	\caption{An overview of the  architecture; adapted from \cite{Borgeaud2021}.}
	\label{fig:mqcnn-cr}
\end{figure}

The overall architecture of our model, MQRetNN, is depicted in Figure \ref{fig:mqcnn-cr}, and we adopt a similar mechanism to incorporate cross-entity information for NLP tasks as shown in Figure 2 of \cite{Borgeaud2021}.

\subsection{Retreival Mechanisms}
\label{sec:method:retrieval}
In this paper, retrieval mechanisms play a key role in scaling up the model to large datasets. In particular, we denote the offline database as $\Hb^0 := \{ h^0_{i,t} | \forall i, t \}$ which consists of the encoded contexts produced by a pre-trained MQCNN encoder. The retrieval calculations are only based on $\Hb^0$ to determine which entities for each example to attend over during training.





\begin{figure}
	\centering
	\includegraphics[width=.4\textwidth]{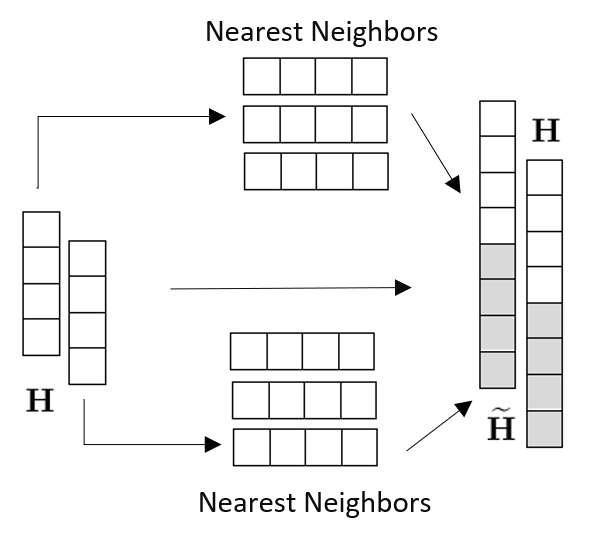}
	\hfill
	\includegraphics[width=.45\textwidth]{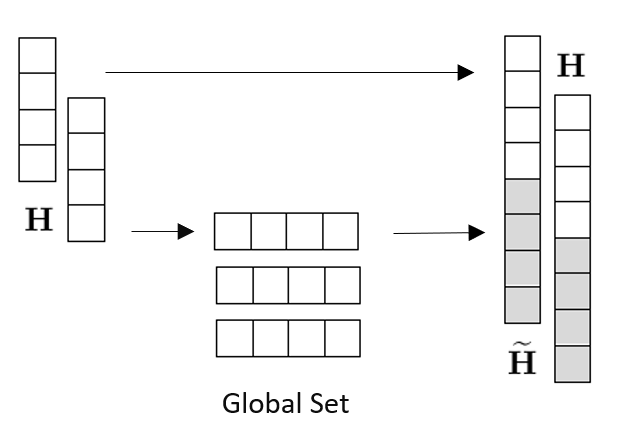}
	\caption{The left diagram depicts nearest neighbor retrieval, the right diagram depicts retrieval using a shared global set. The output of the retrieval step is concatenated to the input embedding vector.}
	\label{fig:cross-att}
\end{figure}

Broadly, we consider two types of retrieval mechanisms:
\begin{enumerate}
	\item Entity-specific retrieval of relevant entities defined as nearest neighbors.
	\item A shared set of entities from the population that are ``maximally relevant'' and used to produce the cross entity context for each entity.
\end{enumerate}
See Figure \ref{fig:cross-att} for a visualization of the two different approaches.

\paragraph*{Entity-Specific Nearest Neighbors}
For each entity $i$, we consider searching for the nearest nearest neighbors in our offline database. This can be formulated as:

\begin{equation}
\label{eq:simRet}
\argmax_{S; |S| = K} \sum_{j\in S, j\neq i} f(h^{0}_{i,t}, h^{0}_{j,t}), \quad \forall i, t.
\end{equation}

Here we find a set of $K$ elements that maximize some similarity metric between example $i$ and elements in $S$, and we search for such set at each time step $t$; that is, we find a time-specific set of $k$ nearest neighbors. We can take $f(\cdot, \cdot)$ to be any similarity metric, and in this paper we consider the Pearson correlation -- which is essentially equivalent to the dot-product attention -- and is computed as
\[
f(h^{0}_{i,t}, h^{0}_{j,t}) = \frac{ <h^{0,c}_{i,t}, h^{0,c}_{j,t}>}{\|h^{0,c}_{i,t}\| \| h^{0,c}_{j,t}\| }
\]
where $h^{0,c}_{j,t}$ denotes the centered version of $h^{0}_{j, t}$.

\paragraph*{Global Set via Submodular Maximization}
Denote by $L(\theta;S)$ the loss in Equation \eqref{eqn:train_loss} evaluated for a model with parameters $\theta$ and set of $S$ of entities to attend over, and let  $\cA$ denote the set of all entities. We would like to select a set $S$ of size $K$ such that
\[
\argmin_{S\subseteq \cA : |S| \leq K} \min_\theta L(\theta; S).
\]
Solving the outer minimization above is not tractable, so instead we consider using a submodular proxy objective to select the set $S$. Specifically we formulate this as follows for each time $t$:

\begin{equation}
\label{eq:subSum}
\argmax_{S; |S| < k} \sum_{i\in \cA} \max_{j\in S, j\neq i} f(h^{0}_{i,t}, h^{0}_{j,t}).
\end{equation}

Here we use the same similarity metric $f$ as in Equation \eqref{eq:simRet}. In general, the form of Equation \ref{eq:subSum} is referred to as the \emph{Facility Location} problem in \cite{Krause2014}, and it is a classic example of optimizing submodular function.

\subsection*{Time-Specific vs. Time-Agnostic Retrieval}
Equation \eqref{eq:simRet} and \eqref{eq:subSum} require the retrieval calculation to be carried out at each time step, both for model training and inference. The advantage is that the retrieval process can then be adaptive to each time step (e.g., nearest neighbors can be different for each time step) and is performed on-the-fly at the inference time. But it can be computationally expensive. Alternatively we propose using $v^{0}_i := \sum^{T}_{t=1} h^{0}_{i,t}$ instead of $h^{0}_{i,t}$ as follows
\begin{align}
& \argmax_{S; |S| = k} \sum_{j\in S, j\neq i} f(v^{0}_i, v^{0}_j) \label{eq:sim2}                 \\
& \argmax_{S; |S| < k} \sum_{i\in \cA} \max_{j\in S, j\neq i} f(v^{0}_i, v^{0}_j) \label{eq:sub2}
\end{align}
i.e., we define the retrieved set $S$ to be \emph{time-agnostic} by considering all time steps in the training window rather than \emph{time-specific} as done previously. In this case, we use the exact same set of entities (but with different contexts for the test period) for model inference and no more retrieval calculation is needed. Note that this does not lead to any information leakage as no computation is done on the test set. We compare the performance of these two types of retrieval mechanisms in the next section.

\section{Results}
\label{sec:results}
In this section we evaluate on a large demand forecasting dataset using two different experimental setups. The dataset comes from a large e-commerce retailer and includes time series features such as demand, promotions, holidays and detail page views as well as static metadata features such as catalog information. Similar datasets with the same set of features but generated in different time windows have been used in \cite{Wen2017, Eisenach2020}. Here we have four years (2015-2019) of data for approximately over 2 million products. The task is to forecast the 50th and 90th quantiles of demand for each of the next 52 weeks at each forecast creation time $t$.

Each model is trained using up to 8 NVIDIA V100 Tensor Core GPUs, on three years of data (2015-2018) and one year is held out for evaluation (2019). In the ``small scale'' setup, we consider only 10,000 different products (entities) so that we can directly attend over a representation of all products rather than use any retrieval method. In the ``large scale'' setup, we have too many to directly attend over all of them simultaneously. Instead, we demonstrate our model can scale up to the entire dataset using retrieval methods, which we ablate and compare the resulting model performance.

\subsection{Small Scale}
In this experiment we choose the 10K products with the largest total units sold during the training period, and we compare four different architectures:
\begin{itemize}
	\item MQCNN: baseline MQCNN model
	\item MQCNN-L: MQCNN with the increased model capacity
	\item MQRet-Full: MQCNN with cross entity context $\tilde{h}_{i,t}$ produced by attending the frozen context across \emph{all other} entities at time $t$ (i.e. from the database $\Hb^0$).
	\item MQRet-Random: Same as above, but where all $\tilde{h}_{i,t}$ are randomly generated.
\end{itemize}

By comparing MQRet-Full with other models, we can better understand how much improvement is possible by augmenting the model with the cross-entity context generated from the entire population. We include two ablations to confirm that the improvement in performance is due to extracting useful cross-entity information. In particular, for testing whether increasing model capacity can lead to performance gain, we consider MQCNN-L which expands MQCNN's capacity by increasing the number of filters of the CNN layer, so that MQRet-Full and MQCNN-L have the same number of parameters. We also consider MQRet-Random, which has the same architecture as MQRet-Full but with randomly generated (non-informative) contexts. We train each model to 100 epochs using batch size of 256, and optimize using ADAM \citep{Kingma2014}. Table \ref{tab:small-exp-param} gives the number of parameters in each trained model.
\begin{table}[!ht]
	\caption{The number of parameters used in the four different architectures of the small scale experiment.}
	\label{tab:small-exp-param}
	\begin{center}
		\begin{tabular}{lc}
			\toprule
			Model        & Number of Parameters \\
			\midrule
			MQCNN        & $0.86 \times 10^6$   \\
			MQCNN-L      & $1.22 \times 10^6 $  \\
			MQRet-Random & $1.21 \times 10^6 $  \\
			MQRet-Full   & $1.21 \times 10^6 $  \\
			\bottomrule
		\end{tabular}
	\end{center}
\end{table}

\begin{table}[!ht]
	\caption{Experiment results on 10K products. All results are rescaled so they are relative improvements over the baseline MQCNN model, lower is better.}
	\label{tab:small-exp}
	\begin{center}
		\begin{tabular}{lccc}
			\toprule
			Model        & P50   & P90   & Overall \\
			\midrule
			MQCNN        & 1.000 & 1.000 & 1.000   \\
			MQCNN-L      & 0.990 & 0.996 & 0.993   \\
			MQRet-Random & 1.008 & 1.007 & 1.008   \\
			MQRet-Full   & 0.968 & 0.978 & 0.973   \\
			\bottomrule
		\end{tabular}
	\end{center}
\end{table}

Table \ref{tab:small-exp} shows the (rescaled) quantile loss results (P50, P90 and overall) for the four models described above. We calculate these results based on three different runs of each model and average the performance metrics. As expected, we observe no accuracy gains from MQRet-Random, as there is no signal to extract. MQCNN-L yields very slight improvement by simply increasing the model capacity. By contrast, MQRet-Full brings relatively substantial improvements in overall performance, improving P50 by 3.2\% and P90 by 2.2\%. Thus, the model seems to be extracting useful signal from other entities.

\subsection{Large Scale}
For this experiment, we use the whole dataset of over 2 million products. The training and test split is kept the same as in the first experiment.

We evaluate the  architecture in Figure \ref{fig:mqcnn-cr} with both retrieval mechanisms described previously, and consider both time-specific and time-agnostic variants. For the nearest neighbor method, we use FAISS \cite{faiss2017}, an open source library for fast nearest neighbor retrieval in high dimensional spaces, and we set $k = 10$. For the submodular method, we use Apricot \cite{apricot2019} which provides efficient submodular optimization tools. In this case, we choose $k = 10000$ for the size of the global set. We selected these values for $K$ to maximize utilization of available GPU memory.

Overall, we consider the following MQRet model variants:
\begin{itemize}
	\item MQCNN: baseline MQCNN model
	\item MQRet-KNN:  with time-agnostic, nearest neighbor retrieval.
	\item MQRet-Subm:  with time-agnostic, submodular retrieval.
	\item MQRet-KNN-t:  with time-specific, nearest neighbor retrieval.
	\item MQRet-Subm-t:  with time-specific, submodular retrieval.
\end{itemize}

\begin{table}[!ht]
	\caption{Experiment results on the whole dataset. Results are rescaled so they are relative improvements over the baseline MQCNN model, lower is better.}
	\label{tab:large-exp}
	\begin{center}
		\begin{tabular}{llccc}
			\toprule
			& Model        & P50   & P90   & Overall \\
			\cmidrule{2-5}
			\parbox[t]{2mm}{\multirow{5}{*}{\rotatebox[origin=c]{90}{All Horizons}}} & MQCNN        & 1.000 & 1.000 & 1.000   \\
			& MQRet-KNN    & 0.999 & 0.973 & 0.987   \\
			& MQRet-KNN-t  & 0.991 & 0.986 & 0.989   \\
			& MQRet-Subm   & 0.993 & 0.996 & 0.994   \\
			& MQRet-Subm-t & 0.991 & 0.988 & 0.990   \\
			\cmidrule{2-5}
			\parbox[t]{2mm}{\multirow{5}{*}{\rotatebox[origin=c]{90}{$h \leq 10$}}}  & MQCNN        & 1.000 & 1.000 & 1.000   \\
			& MQRet-KNN    & 0.995 & 0.971 & 0.984   \\
			& MQRet-KNN-t  & 0.986 & 0.993 & 0.989   \\
			& MQRet-Subm   & 0.983 & 0.989 & 0.986   \\
			& MQRet-Subm-t & 0.985 & 0.991 & 0.985   \\
			\bottomrule
		\end{tabular}
	\end{center}
\end{table}

We train each model for 100 epochs with a batch size of 512. Test results are summarized in Table \ref{tab:large-exp}. We include the model performance aggregated across all horizons (52 weeks) as well as for horizons $h \leq 10$. We observe that all MQRet variants improve the overall performance by around 1\% but the gains are smaller than the full cross-entity attention in Table \ref{tab:small-exp}. Larger performance improvements are observed for all models when aggregated over shorter horizons. The performance of time-specific models are generally similar to that of time-agnostic ones. The best variant -- MQRet-KNN --  improves by 1.3\% over the baseline MQCNN model for all horizons, and by 1.5\% when restricted to only shorter horizons ($h\leq 10$).

\section{Conclusion}

In this paper we demonstrated that incorporating cross-entity information can improve the predictive accuracy of time-series forecasting models. On our target application, we showed approximately a 3\% improvement over the baseline model when we attended over all other entities in the population. The gains on the large scale dataset were smaller -- approximately over 1\% improvement on the baseline. Accordingly, a future directions of interest is training a model that can attend across all entities during each forward pass -- will require model parallelism across multiple machines. Another interesting direction of future inquiry is using pretrained graphs between entities to select the nearest neighbors.

\bibliographystyle{ims_nourl_eprint}
\bibliography{references}

\end{document}